\documentclass[10pt,twocolumn,letterpaper]{article}

\usepackage{iccv}
\usepackage{times}
\usepackage{epsfig}
\usepackage{graphicx}
\usepackage{amsmath}
\usepackage{amssymb}
\usepackage{color}
\usepackage{algorithm}
\usepackage{algorithmic}
\usepackage{caption}
\usepackage[table]{xcolor}
\usepackage{epstopdf}
\usepackage{enumitem}
\DeclareGraphicsExtensions{.pdf,.png,.jpg}
\newcommand{\figref}[1]{Fig.~\ref{#1}}

\newcommand{\eqnref}[1]{Eq.~(\ref{#1})}
\newcommand{\secref}[1]{Sec.~\ref{#1}}

\renewcommand{\ie}{\textit{i.e.}}
\renewcommand{\eg}{\textit{e.g.}}

\usepackage[pagebackref=true,breaklinks=true,letterpaper=true,colorlinks,bookmarks=false]{hyperref}

\iccvfinalcopy 

\def\iccvPaperID{1241} 

\ificcvfinal\fi
\begin{document}

\title{Weakly- and Self-Supervised Learning for Content-Aware Deep Image Retargeting}

\author{Donghyeon Cho\\
KAIST\\
\and
Jinsun Park\\
KAIST\\
\and
Tae-Hyun Oh\\
KAIST\\
\and
Yu-Wing Tai\\
Tencent Youtu\\
\and
In So Kweon\\
KAIST\\
\and
{\tt\small $\lbrace$cdh12242, zzangjinsun, thoh.kaist.ac.kr, yuwing$\rbrace$@gmail.com, iskweon@kaist.ac.kr}
}

\maketitle
\thispagestyle{empty}

\begin{abstract}
This paper proposes a weakly- and self-supervised deep convolutional neural network (WSSDCNN) for content-aware image retargeting. Our network takes a source image and a target aspect ratio, and then directly outputs a retargeted image. Retargeting is performed through a shift map, which is a pixel-wise mapping from the source to the target grid. Our method implicitly learns an attention map, which leads to a content-aware shift map for image retargeting. As a result, discriminative parts in an image are preserved, while background regions are adjusted seamlessly. In the training phase, pairs of an image and its image-level annotation are used to compute content and structure losses. We demonstrate the effectiveness of our proposed method for a retargeting application with insightful analyses.
\end{abstract}

\section{Introduction}

\begin{figure}
\begin{center}
\includegraphics[height=0.53\linewidth]{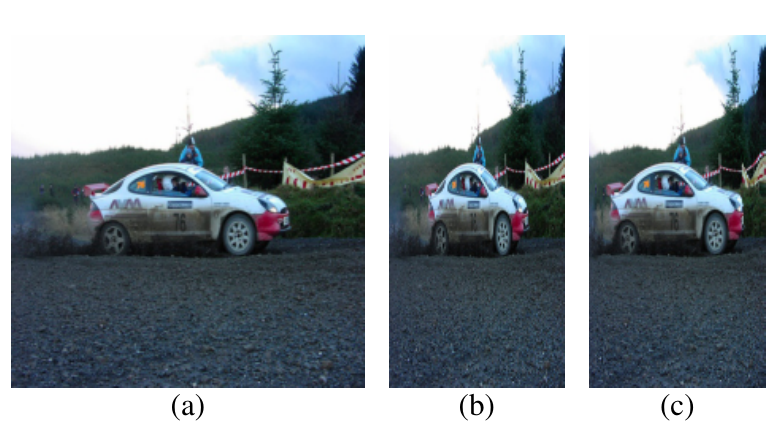}
\end{center}
\vspace{-0.2in}
\caption{Content-aware retargeting examples. (a) An input image. (b) Linear scaling. (c) Content-aware retargeting (\texttt{OURS}). Our method tends to preserve important parts of the given image well.}
\label{fig:Intro}
\vspace{-0.2in}
\end{figure}

Cameras and display devices are designed depending on diverse targets of customers' needs, and hence the resolution and aspect ratio of each module are different.
Considering a full screen display scenario of an image, the original image may not perfectly fit the display in full screen, due to the different aspect ratios between the display device and the image. It may rather introduce clipping, stretching or shrinking, as shown in~\figref{fig:Intro}-(b).

Image retargeting research has been actively carried out in computer vision and image processing. Image retargeting techniques adjust the aspect ratio (or size) of an image to fit the target aspect ratio, while not discarding important content in an image. \textit{Content-aware image retargeting} aims to preserve important content as much as possible, and  various methods have been suggested~\cite{Liu2005AIR,WOLFGCO2007ICCV,Wang2008TOG,Zhang2009CGF,Guo2009TM,Jin2010TVC,Avidan2007TOG,Rubinstein2008TOG,Shamir2009Commun,Han2010TVC,Michael2011SPL,Dong16TVCG}. 

In traditional content-aware approaches, salient region information is widely used.
The saliency is conventionally generated from edge map or hand-crafted features, which are restricted by fixed design principles and limit its generality.
In recent years, deep learning studies have been actively conducted to extract high-level semantic features~\cite{ILSVRC15,Simonyan14c,Googlenet,Girshick14cvpr,girshick15ICCV15} of an image. Furthermore, research on deep learning has shown excellent performance in low-level image processing tasks~\cite{NIPS12Xie,NIPS14Xu,Eigen2013iccv,Ren2015nips,ECCV14Chao,Xu2015icml,park17CVPR,lee16CVPR}.



Motivated by this, in this paper, we propose a content-aware image retargeting method using high-level semantic information through deep learning. In order to resize input images within a network, we introduce a shift layer that maps each pixel from the source to the target grid. We demonstrate that end-to-end training is possible through the shift layer, and retargeted images are directly produced using input images with target aspect ratios. The spatial semantic  information is learned from image-level annotations, and passed to the shift layer. Image-level annotations are used to compute content loss on the retargeted images. Moreover, input images are used for structure loss to suppress unwanted visual effects after retargeting.
In summary, this paper provides the following contributions:

\begin{enumerate}[leftmargin=4mm]
  \item We introduce a weakly- and self-supervised learning for content-aware deep image retargeting. 
  We utilize images and its corresponding image-level annotations for structure and content loss computations, respectively. They do not require much human effort for output labels to supervise network.
  \item Our proposed network takes a source image and a target aspect ratio as input, and then directly produces a retargeted image in a shot. Therefore, end-to-end training is possible and its test time is also fast. 
  To our knowledge, this work is the first attempt to apply deep learning to the image retargeting application. 
  \item We design a shift layer that maps each pixel from the source to the target grid. Our method implicitly learns semantic attention information, and passes it to the shift map. 
\end{enumerate}


\section{Related Works}
In this section, we review previous studies related to this paper. In particular, we review recent image processing research based on deep CNN models, and then review image retargeting approaches.


\paragraph{Deep CNN Model for Low-Level Vision Tasks}
Recently, there have been many works related to CNN for computer vision problems. 
We introduce a few insightful works that are relevant to image processing problems, and all of these pioneer works have demonstrated successful results: image denoising~\cite{NIPS08Jain,NIPS12Xie}, image artifact removal~\cite{ICCV13Eigen,Chao15ICCV}, super resolution~\cite{ECCV14Chao,KimLL15a,KimLL15b}, deblurring~\cite{NIPS14Xu,CVPR15Sun}, colorization~\cite{ICCV15Cheng,arXiv16zhang}, image inpainting~\cite{NIPS12Xie,pathak16CVPR,Raymond16c}, and image matting~\cite{Shen16ECCV, Cho16ECCV}. Interestingly, researchers have shed little light on image retargeting based on a deep CNN approach. We sketch the advances of image retargeting works in a categorical manner as follows.


\paragraph{Image Retargeting}
Approaches for image retargeting can be categorized in a variety of ways, but we roughly classify them into seam carving based and warping based methods. The \textit{seam carving based} methods~\cite{Avidan2007TOG,Rubinstein2008TOG,Shamir2009Commun,Han2010TVC,Michael2011SPL} change the aspect ratio of an image by repeatedly removing or insetting seams at unimportant areas. Avida \textit{et al.}~\cite{Avidan2007TOG} introduce the concept of seam carving and solve it using dynamic programming. Rubinstein \textit{et al.}~\cite{Rubinstein2008TOG} later apply seam carving in 3D volume for video retargeting. A representation of multisize media is defined by Shamir \textit{et al.}~\cite{Shamir2009Commun} to have continuous resizing ability in real time. Han \textit{et al.}~\cite{Han2010TVC} find multiple seams simultaneously with region smoothness and seam shape prior. Frankovich and Wong~\cite{Michael2011SPL} propose an enhanced seam carving method by incorporating energy gradient information into optimization framework.

The \textit{warping based} approaches ~\cite{Liu2005AIR,WOLFGCO2007ICCV,Wang2008TOG,Zhang2009CGF,Guo2009TM,Jin2010TVC} continuously transform an input image into an image of a target size. In Liu and Gleicher~\cite{Liu2005AIR}, a non-linear image warping is used to emphasize important parts of an image. Wolf \textit{et al.}~\cite{WOLFGCO2007ICCV} try to reduce distortion by shrinking less important pixels and preserving important regions. A scale-and-stretch warping method is proposed by Wang \textit{et al.}~\cite{Wang2008TOG}. Their method iteratively computes optimal scaling factors for local regions and updates a warped image assisted by edge and saliency map. 
Guo \textit{et al.}~\cite{Guo2009TM} suggest a mesh representation based on image structures to preserve the shape of an input image. An interactive content-aware retargeting is proposed by Jin \textit{et al.}~\cite{Jin2010TVC}. They formulate retargeting using a sparse linear system based on a triangular mesh, and then solve it via quadratic optimization.

Studies that have been conducted thus far have used salient region information of an image for content-aware image retargeting. However, in most seam carving or warping based methods, previous studies have used hand-crafted features such as edge maps to find prominent areas of an image. In this paper, we apply deep CNN for image retargeting to reflect more semantic information by utilizing many images for training. Our method directly produces a retargeted image from an input image with a target size, and it does not require iterative processes. As far as we are aware, this is the first work that seriously addresses the content-aware image retargeting through the weakly- and self-supervised deep CNN model.


\begin{figure*}[t]
\begin{center}
   \includegraphics[width=0.99\linewidth]{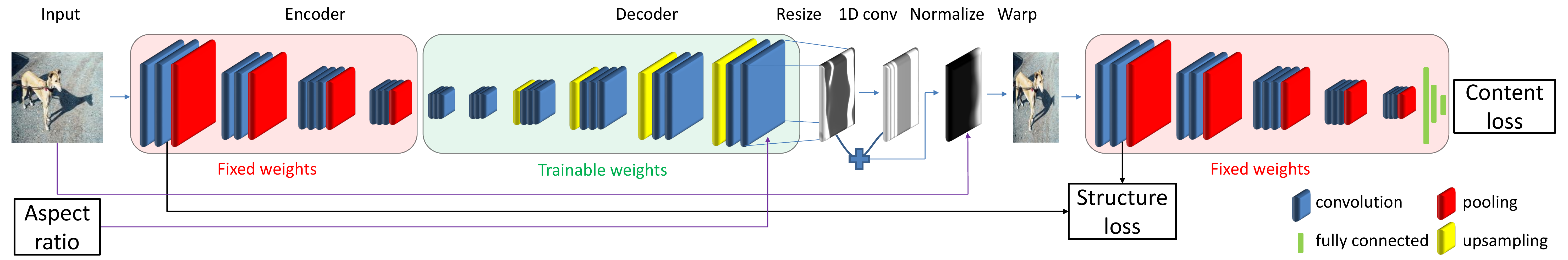}
\end{center}
\vspace{-0.2in}
   \caption{Overall architecture of the proposed network for image retargeting.}   
   \label{fig:overall}
   \vspace{-0.2in}
\end{figure*}

\section{Architecture}
\label{sec:Architecture}
In this section, we introduce our WSSDCNN model for image retargeting. The overall architecture consists mainly of two parts: an encoder-decoder~(\secref{sec:en_de}) and a shift layer~(\secref{sec:shift}) as illustrated in~\figref{fig:overall}. The aspect ratio can be adjusted by changing the width or height of an image. For simplicity, we describe only the case of reducing the width of images, but one can understand the other cases analogously with simple modification, \eg, height adjustment can be done by rotating an image by 90 degrees.



\subsection{Encoder-Decoder Model}
\label{sec:en_de}
The most important information for the content-aware image retargeting is semantic information for each object in the scene. Recently, some of previous works~\cite{krizhevsky2012imagenet, zeiler2010deconvolutional} have discovered that activations from high-level layers tend to encode semantic information (i.e. high-level information). Based on this observation, in order to utilize semantic information, the encoder of our network takes an input image and extract high-level features containing semantic information. The decoder generates an attention map using high-level features from the encoder. We adopt pre-trained VGG16~\cite{Simonyan14c} and inversely symmetric VGG16 architectures for the encoder and decoder, respectively. In addition, we remove fully connected layers of VGG16 and replace ReLU layers in the decoder with ELU~\cite{clevert2015fast} layers.


\subsection{Shift Layer}
\label{sec:shift}
The shift map defines to what extent a pixel should be shifted from the input of size $W \times H$ to specified output grids of size $W' \times H$ for each pixel. 
The relationship between the input and output images via shift map is   
\begin{eqnarray}
O(x,y) = I(x+S(x,y),y),
\label{eq:shift_map}
\end{eqnarray}
where $I$ and $O$ denote input and output images and $(x,y)$ are spatial coordinates of the output grid, respectively. \hbox{$S \in [0, W-W']^{W' \times H}$} is the shift map for the output grid.


We utilize an attention map from the decoder, whereby the shift map acts in a  semantic aware manner.
First, we resize the attention map to target size. 
\begin{eqnarray}
A_{r}= \mathrm{resize}(A_{d},R),
\label{eq:S_r}
\end{eqnarray}
where $A_{r}$, $A_{d}$, and $R$ denote a resized attention map, an output from decoder, and a target aspect ratio, respectively. We then feed it into 1D duplicate convolution and cumulative normalization layers, which are described as follows.

\begin{figure}[t]
\begin{center}
   \includegraphics[width=0.99\linewidth]{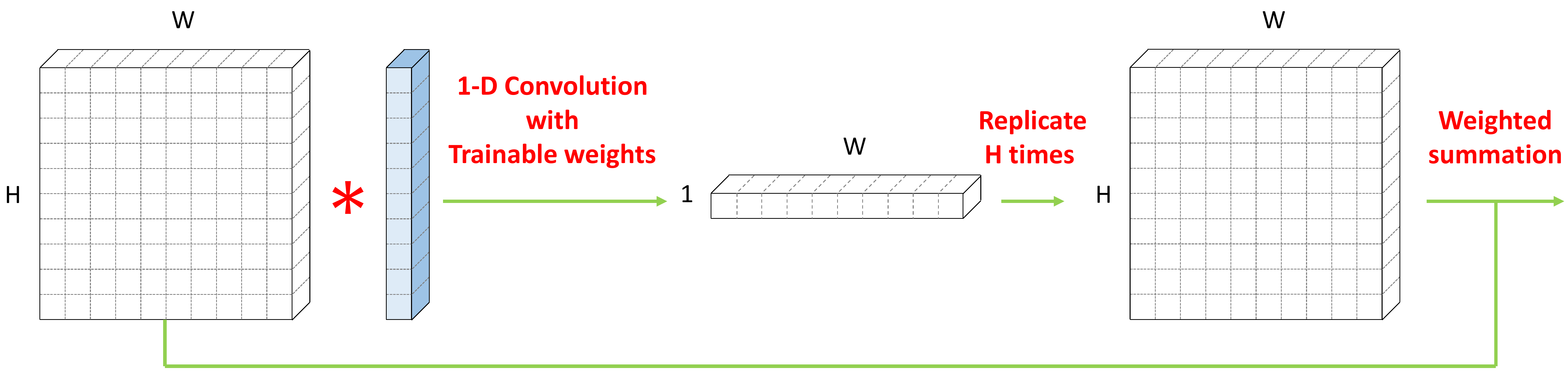}
\end{center}
\vspace{-0.2in}
   \caption{Illustration of 1D duplicate convolution. It reduces visual artifacts after retargeting effectively as shown in \figref{fig:distort}.}
   \label{fig:1d_conv}
   \vspace{-0.2in}
\end{figure}

\paragraph{1D Duplicate Convolution}
\label{sec:1d_conv}
An attention map has rough localization information about discriminative parts in an image. By transforming semantic driven location information through the shift map, semantically important parts in an image are preserved while the scales of background regions are adjusted. However, in order to maintain the overall shape of an image, pixels in similar columns should have similar shift values. Therefore, we constrain the shape of an attention response to be uniform along the column axis using 1D duplicate convolution layer:
\begin{eqnarray}
A_{1D}= \mathrm{duplication}(\mathrm{conv}_{1D}(A_{r},H),H),
\label{eq:1d_conv}
\end{eqnarray}
where $\mathrm{conv}_{1D}(\cdot,H)$ is a convolution with a $H$ dimensional column vector without padding, and $\mathrm{duplication}(\cdot,H)$ repeats one dimensional vector $H$ times as shown in~\figref{fig:1d_conv}. All weights of $\mathrm{conv}_{1D}(\cdot,H)$ are learned by training.

Since $A_{1D}$ is restricted to be just column-wise map, to handle residual cases, we use the combination of $A_{r}$ and $A_{1D}$ as a final attention map. 
\begin{eqnarray}
A= \lambda A_{r} +  A_{1D},
\label{eq:sum_shift}
\end{eqnarray}
where $\lambda$ is a balancing parameter for $A_{r}$ and $A_{1D}$. 
The rationale behind this composition is rather emblematic. By restricting the structure of attention map $A_{1D}$, it captures the majority of the attention while making the final map vertically structured, whereas the role of $A_r$ is reduced to draw residual attention, which might be necessary. If we directly use an attention map $A_{r}$ from the decoder, background areas are twisted around discriminative objects, as shown in~\figref{fig:distort}-(b).

\paragraph{Cumulative Normalization}
\label{sec:cumulative}
In order to recover the final mapping from the input grid to the target grid, we convert $A$ into shift map $S$. Since $S$ should be monotonically increased along the spatial axes, we perform cumulative normalization to constrain it.
 \begin{eqnarray}
S(x,y) =  \alpha \cdot \frac{\sum_{x' \leq x} A(x',y)}{\sum_{x} A(x,y)},
\label{eq:cum_normalization}
\end{eqnarray}
where $\alpha$ is $|W-W'|$. From the shift map $S(x,y)$, we retarget an input image to the target grid using \eqnref{eq:shift_map}.

\paragraph{Image Warping}
\label{sec:warping}
Finally, by~\eqnref{eq:shift_map}, an input image is warped into an image of a target size. Because the shift map has subpixel precision, linear interpolation is performed using four neighboring pixels. As mentioned in~\cite{Garg16ECCV}, applying a differentiable loss to a linearly warped image is also differentiable. 
To further proceed, when a converted image is passed to a pre-trained network to compute content and structure losses, which are described in the following section, zero padding is performed to fit the input size of the pretrained model.

\section{Losses}
\label{sec:Losses}
In order to train the shift map, we utilize two types of supervisions. Image-level annotations and input images themselves are used for content and structure loss calculation, respectively. 

\subsection{Content Loss}
If the main content in an image is well preserved after retargeting, its retargeted image should have a similar outcome of classification using the original image. 
Taking this point into consideration, we define content loss. Unlike content loss used in~\cite{Gatys2016CVPR,Johnson16c} to represent reconstruction errors that measure activation differences of the target and the prediction, we use a class score loss for the classification task as content loss in this paper. The content loss is calculated on the retargeted image, and defined as per-class cross entropy loss on sigmoid outputs from a classifier.
\begin{align}
E_{c} = -\frac{1}{CN}\sum_{k=1}^{C}\sum_{i=1}^{N}\left [ l_{ki}\log \hat{l_{ki}} + (1-l_{ki})\log (1-\hat{l_{ki}}) \right ],
\label{eq:class_loss}
\end{align}
where $C$ and $N$ are the number of classes and samples. $l_{ki}$ and $\hat{l_{ki}}$ are the ground truth label and sigmoid output, respectively. We use VGG16~\cite{Simonyan14c} as a classifier model, and all weights are initialized by pre-trained values, and fixed during both training and test phases.

In order for classification to work well after retargeting, discriminative parts of an image have to be well preserved. Therefore, by content loss, the shift map is trained so as to implicitly contain information about prominent areas of an image.


\subsection{Structure Loss}

While preserving the most important parts of the given image, we want to suppress unnatural visual artifacts such as distortion. In order to make the retargeted image have a similar structure to the original image, we utilize the input image as a supervision for structural similarity. We design structure loss to make the neighborhood of each pixel in the retargeted image to be as similar as possible to the neighborhood of each corresponding pixel in the input image. Since we can infer correspondences between the input image and the retargeted image from the shift map, we can measure the corresponding patch similarity of the input and output. One of the simplest approaches for the similarity measure is pixel-wise 4-neighbor comparison between input and output pixels. However, pixel-wise comparison easily fails with image noise, contrast change and patch misalignment.

As reported in ~\cite{krizhevsky2012imagenet, zeiler2010deconvolutional}, activations from the first few convolutional layers of CNN provide low-level structural information. These activations are robust to image noise, contrast change, and misalignment due to convolution operations.
Therefore, we do not utilize pixel-wise difference, but utilize activations from $conv1$ of VGG16 for the computation of structural similarity as follows:
\begin{eqnarray}
E_{s} = \sum_{j=1}^{2} \sum_{x,y} [\mathbf{F_j}(O(x,y)) - \mathbf{F_j}(I(x+S(x,y),y))],
\label{eq:structure_loss}
\end{eqnarray}
where $\mathbf{F_j}$ are function of $conv1_j$ in VGG16. Through $conv1_1$ and $conv1_2$, the receptive field of each pixel covers a $5 \times 5$ neighborhood.

\begin{figure}
\begin{center}
\def\arraystretch{0.5}
\begin{tabular}{@{}c@{\hskip 0.01\linewidth}c@{\hskip 0.01\linewidth}c}
\includegraphics[height=0.4\linewidth]{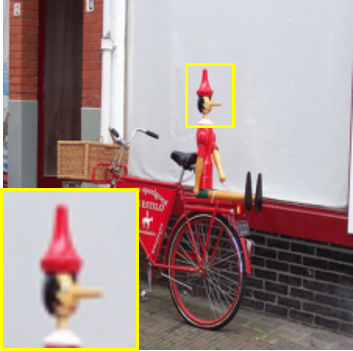} &
\includegraphics[height=0.4\linewidth]{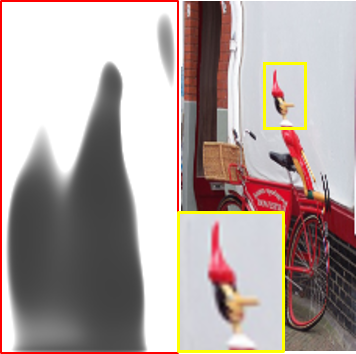}\\
{\small (a)} & {\small (b)}\\
\includegraphics[height=0.4\linewidth]{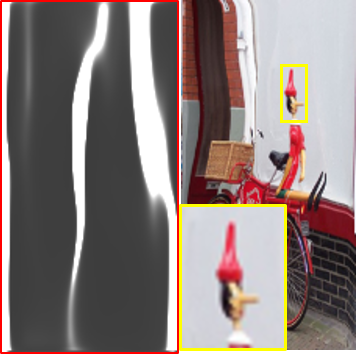} &
\includegraphics[height=0.4\linewidth]{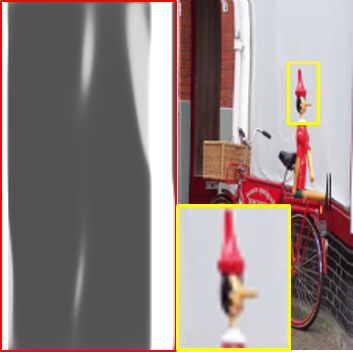} \\
{\small (c)} & {\small (d)}
\end{tabular}
\end{center}
\vspace{-0.2in}
\caption{(a) An input image. (b-d) A attention map and a regtargeted image. (b) Content loss. (c) Content loss + structure loss. (d) Content loss + structure loss + 1D convolution.}
\label{fig:distort}
\vspace{-0.2in}
\end{figure}

The network trained with structure loss makes pixels in similar columns of an attention map to have similar values, as shown in~\figref{fig:distort}-(c). This is desirable in the 1D duplicate convolution layer, and, as a result, artifacts of a retargeted image are significantly reduced compared to the case using only content loss. Also, combining the structure loss with the 1D duplicate convolution layer produces better results, as in~\figref{fig:distort}-(d).
In short, content loss plays a role of preserving salient objects, and the structure loss and the 1D duplicate convolution layer reduce the artifact of retargeted images.

\section{Experiments}
In this section, we first describe implementation details. Qualitative comparisons of deep content-aware image retargeting to previous works are also shown. After that, we analyze the role and effect of the shift map of our algorithm.


\subsection{Training}
We train our proposed network using Pascal VOC 2007 dataset~\cite{pascal-voc-2007} with only image-level annotations. First of all, we train the VGG16 model using VOC data, and use the trained weights for initialization of the encoder and classifier parts. They are fixed during the training of the other parts. The other parts of the network are trained with content and structure losses. Input images are resized to $224 \times 224$, and input aspect ratios are randomly generated for each batch within $ \frac{224}{4} \sim  \frac{224}{2}$. It takes around $1\sim2$ days for 300 epochs on a machine with a GTX 1080 GPU and an Intel i7 3.4GHz CPU. The learning rate, momentum and batch size are set to $10^{-−5}$, 0.9, and 16, respectively. A forward pass takes around $0.5\sim1$ seconds to process an image with a resolution of $224 \times 224$ pixels. Except for the classifier used in training, the proposed network is fully convolutional, and thus an image of arbitrary size can be retargeted at the test time. 

\subsection{Results}
\label{sec:results}
\begin{figure*}
\begin{center}
\includegraphics[width=0.9\linewidth]{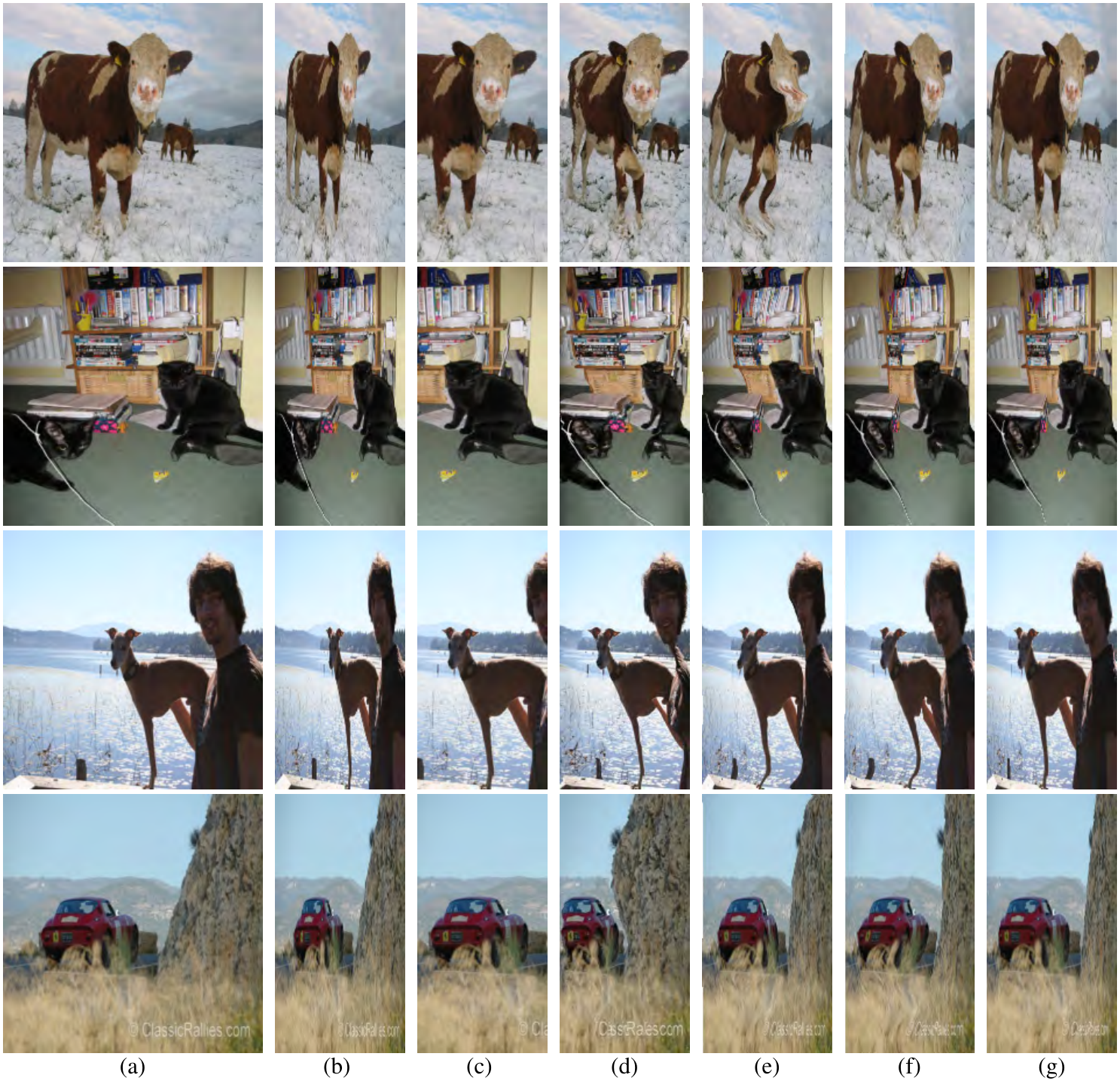}
\end{center}
\vspace{-0.3in}
\caption{Width adjustment. (a) Input images. (B) Linear scaling. (c) Manual Cropping. (d) Seam carving~\cite{Avidan2007TOG}. (e) Results with content loss. (f)  Results with content loss + structure loss. (g) Results with content loss + structure loss + 1D duplicate convolution.}
\label{fig:Width_adjustment}
\end{figure*}

\begin{figure*}
\begin{center}
\includegraphics[width=0.9\linewidth]{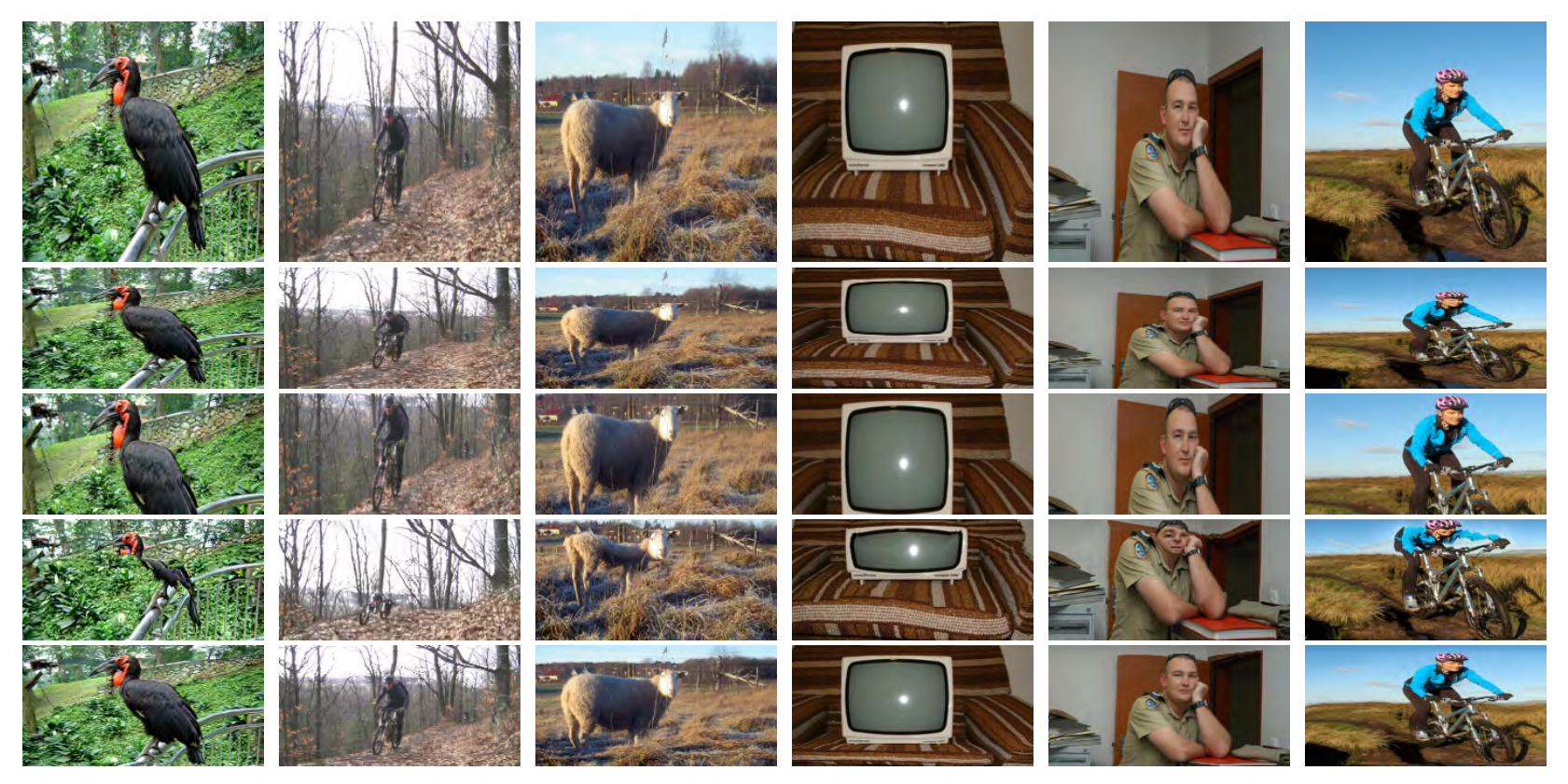}
\end{center}
\vspace{-0.2in}
\caption{Height adjustment. From the top to the bottom: input images, linear scaling, manual cropping, seam carving~\cite{Avidan2007TOG}, and our results (content loss + structure loss + 1D duplicate convolution).}
\label{fig:Height_adjustment}
\end{figure*}

\begin{figure*}
\begin{center}
\includegraphics[width=0.9\linewidth]{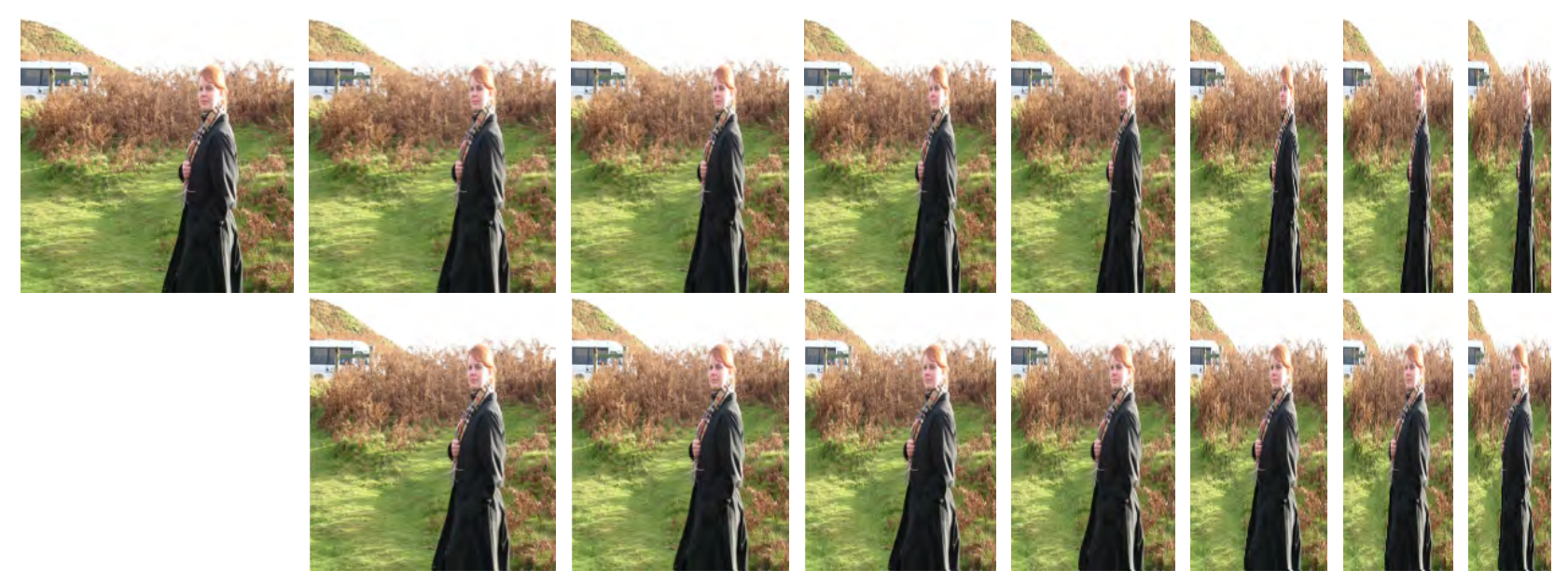}
\end{center}
\vspace{-0.2in}
\caption{Retargeted images according to varying aspect ratio from 0.9 to 0.3 times width. (Top) Linear scaling. (Bottom) Our results (content loss + structure loss + 1D duplicate convolution).}
\label{fig:ratio}
\vspace{-0.1in}
\end{figure*}

\paragraph{VOC Dataset}
\figref{fig:Width_adjustment} shows the width retargeting results on VOC 2007 test images. All results are reduced to half size. We compare our results with linear scaling, manual cropping, and seam carving~\cite{Avidan2007TOG}. In addition, results with only content loss, and results with content loss+structure loss are compared to verify the effects of structure loss and 1D duplicate convolution layer. In linear scaling~(\figref{fig:Width_adjustment}-(b)), the size of the object can be excessively reduced while cropping~(\figref{fig:Width_adjustment}-(c)) takes away important regions of an image. Since seam carving~(\figref{fig:Width_adjustment}-(d)) subtracts the discretized seam, if the reducing factor is large or objects occupy a large portion of an image, important parts are also removed and distorted. Also, as mentioned above, with only content loss~(\figref{fig:Width_adjustment}-(e)), objects are preserved but distorted. Structure loss~(\figref{fig:Width_adjustment}-(f)) suppresses much of the distortion, but there are still wobbles. Through the 1D duplicate convolution layer, however, wobbles are significantly restored, as in~\figref{fig:Width_adjustment}-(g). In addition, height retargeting results are shown in~\figref{fig:Height_adjustment}. Our results are also better than the results of linear scaling, manual cropping and seam carving. Results of progressively reducing the size of an image are shown in~\figref{fig:ratio}. Even though the size of an image can be changed only in discrete pixels, we can make a continuous changing effect if the aspect ratio is densely sampled. 

Interestingly, semantic objects often disappear after seam carving. This is because seam carving depends on low-level features such as edge maps, as well as other existing content-aware image retargeting methods. For instance, in the first and second columns of~\figref{fig:Height_adjustment}, the results of seam carving show that a bird and a bicyclist are severely damaged. This is because seam carving, which relies on low-level features, cannot reasonably infer semantic information in complex background images. This problem can be solved by using a high-level feature using deep learning as shown in results of our proposed method.

\begin{figure*}
\begin{center}
\includegraphics[width=0.9\linewidth]{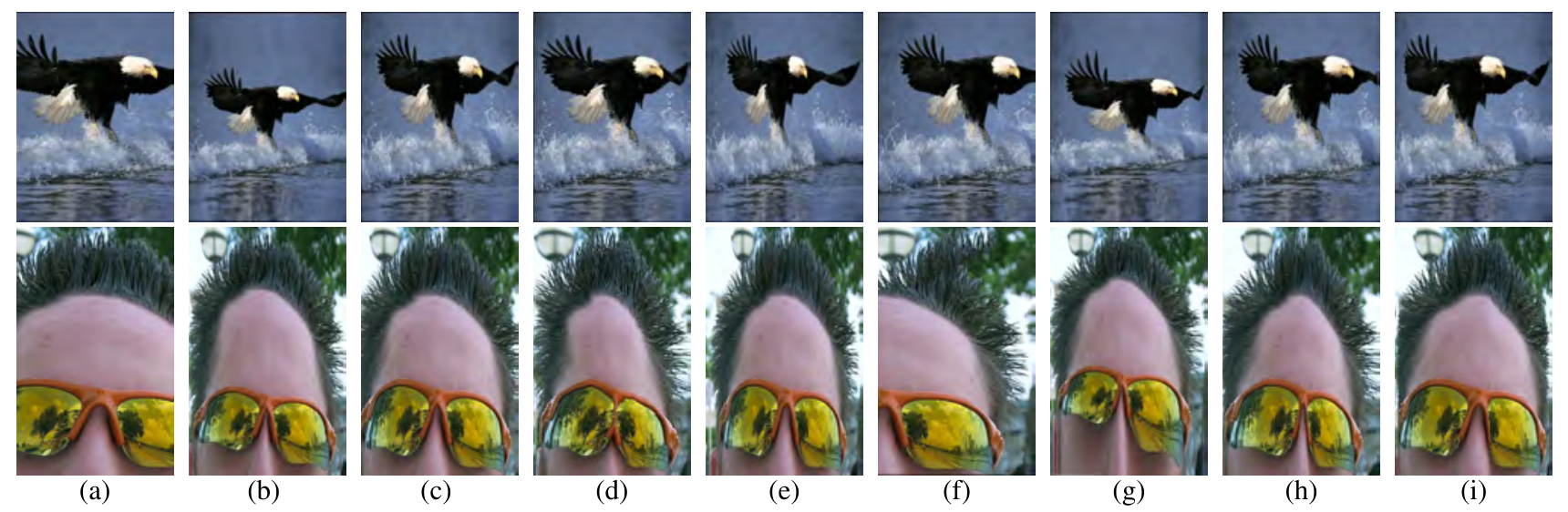}
\end{center}
\vspace{-0.2in}
\caption{Comparisons on benchmark in~\cite{Rubinstein10SIGGRAPHASIA} (a) Manual Cropping. (b) Streaming Video~\cite{Krahenbuhl09TOG}. (c) Multi-operator~\cite{Rubinstein09TOG}. (d) Seam carving~\cite{Rubinstein2008TOG}. (e) Linear scaling. (f) Shift map~\cite{Pritch09ICCV}. (g) Scale-and-Stretch~\cite{Wang2008TOG}. (h) Nonhomogeneous warping~\cite{WOLFGCO2007ICCV}. (i) Our results (content loss + structure loss + 1D duplicate convolution).}
\label{fig:bench}
\vspace{-0.2in}
\end{figure*}

\paragraph{Benchmark Dataset}
In order to compare our method with more previous works, we directly quote published results from a benchmark~\cite{Rubinstein10SIGGRAPHASIA}. 
\figref{fig:bench}-(a-h) shows example results of previous methods~\cite{Krahenbuhl09TOG,Rubinstein09TOG,Rubinstein2008TOG,Pritch09ICCV,Wang2008TOG,WOLFGCO2007ICCV} including cropping and linear scaling. Our results are shown in~\figref{fig:bench}-(i). Although images from~\cite{Rubinstein10SIGGRAPHASIA} do not have exactly the same labels as the VOC dataset, object regions are well found, and our results are competitive with previous methods.

\begin{figure*}
\begin{center}
\includegraphics[width=0.9\linewidth]{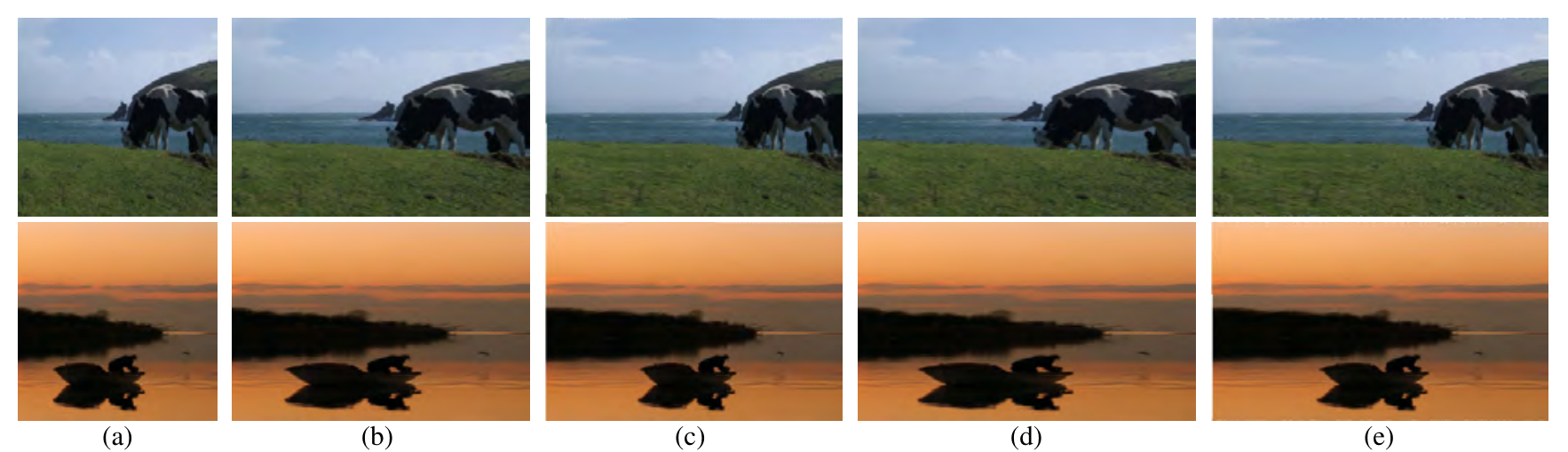} 
\end{center}
\vspace{-0.2in}
\caption{Enlarging examples. (a) Input images. (b-e) $1.5$ and $1.7$ times enlarged images using linear scaling and our method. (b) Linear scaling ($1.5 \times$). (c)  Our method ($1.5 \times$). (d) Linear scaling ($1.7 \times$). (e) Our method ($1.7 \times$).}
\label{fig:enlarge}
\vspace{-0.1in}
\end{figure*}

\paragraph{Image Enlarging}
Although the proposed network is trained to adjust the aspect ratio of an image by reducing the size of an image, it is also possible to appropriately enlarge an image by using an attention map. In the case of image enlarging, since the main objects should be fixed as much as possible while textures and background regions are expanded, an attention map is reversed to map an input to target grid as follow:
\begin{eqnarray}
A^{-1} = \exp(-\frac{A}{\gamma}),
\label{eq:inverse_att}
\end{eqnarray}
where $\gamma$ is a parameter of inversion. After cumulative normalization, the final shift map is obtained. In order to expand an image by $k$, we first obtain an attention map for the task of reducing an input by $k$, and then resize it to the input size. For image enlarging, mapping between an input of size $W \times H$ and an output of size $W' \times H$ images is determined as follows:
\begin{eqnarray}
O'(x'+S(x',y'),y') = I(x',y'),
\label{eq:img_enlarging}
\end{eqnarray}
where $(x',y')$ are spatial coordinates of the input grid. In this case, $S$ ranges from $0$ to $W'-W$. \figref{fig:enlarge} shows examples of image enlarging. Compared to linear scaling, the main objects are much better preserved while background regions are expanded.


\begin{figure}
\begin{center}
\includegraphics[width=0.99\linewidth]{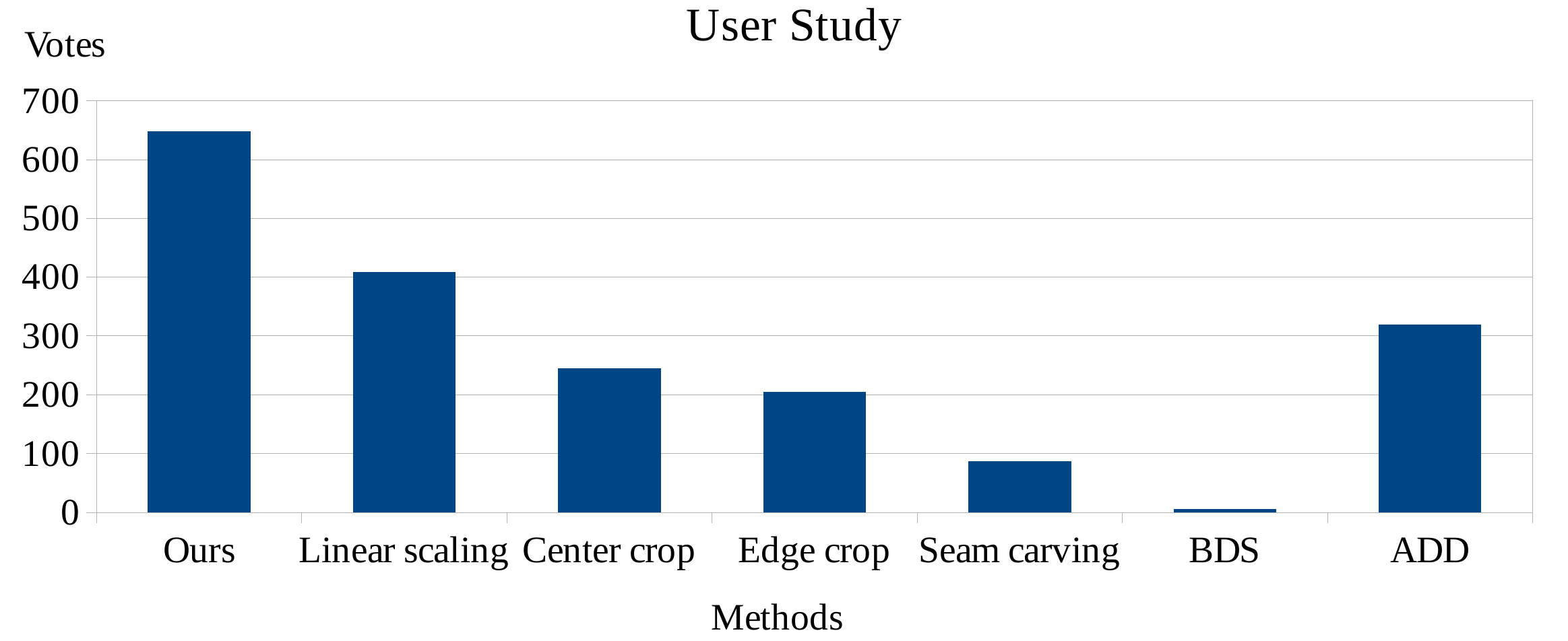}
\end{center}
\vspace{-0.3in}
\caption{User study results.}
\label{fig:user}
\vspace{-0.15in}
\end{figure}

\paragraph{User Study}
In order to evaluate our proposed method in terms of pleasantness to the human eye, we conduct a user study with 32 people from different backgrounds (age range 24-55) using 30 sets of pascal (20) and non-pascal (10) images. For comparison with the proposed method, linear scaling, center crop, edge crop, seam carving, BDS~\cite{Simakov08CVPR}, and AAD~\cite{Panozzo12CGF} methods are used. Each person is guided to select two preferred images over the seven choices. As shown in~\figref{fig:user}, our proposed method receives the highest vote. Also, considering the linear scale method is a moderate baseline, it is qualitatively meaningful that our proposed method receives the most votes.

\subsection{Analysis}
\begin{figure}
\begin{center}
\def\arraystretch{0.5}
\begin{tabular}{@{}c@{\hskip 0.01\linewidth}c@{\hskip 0.01\linewidth}c}
\includegraphics[width=0.5\linewidth]{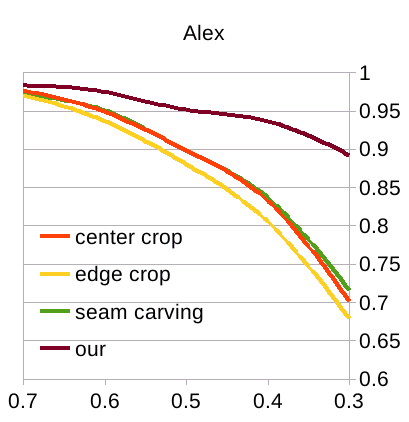} &
\includegraphics[width=0.5\linewidth]{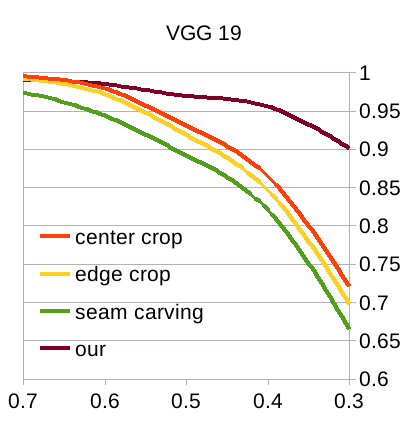}\\
\end{tabular}
\end{center}
\vspace{-0.2in}
\caption{Transition graph of mAP ratio according to the resizing scale.
X-axis: scale ratio of image retargeting, Y-axis: mAP ratio against classification results of original images.}
\vspace{-0.15in}
\label{fig:plot}
\end{figure}

\paragraph{Classification on Retargeted Images}
In order to quantitatively confirm that the main objects are well preserved after retargeting, we experiment on how much the classification accuracy differs before and after retargeting. Experiments are conducted on the VOC 2007 test set. For the purpose of fair comparison, we use pre-trained Alexnet and VGG19 instead of VGG16, so that we can evaluate the generality while avoiding any bias induced by VGG16, which was used in our model. 
Since the VOC dataset has multiple classes in an image, we measure mAP instead of accuracy, and report the ratio with classification results against original images, \ie, $\tfrac{\texttt{mAP}(I_\textrm{retargeted})}{\texttt{mAP}(I_\textrm{original})}$. 
Images with reduced size are linearly interpolated to their original size before classification. Note that we use zero padding when retargeted images are passed to the classifier at training time. We compare the mAP ratio of the proposed method with the retargeting methods using center cropping, edge-based cropping, and seam carving~\cite{Avidan2007TOG}. Edge-based cropping is a way to crop a window containing the largest amount of gradients within the given window size. As shown in~\figref{fig:plot}, the mAP ratio is measured while reducing image sizes from 0.7 to 0.3 times at intervals of 0.1. Our method shows significantly slower decay of performance than the other methods. 
Through these quantitative experiments, we verify that the proposed method can adjust the size of an image while preserving the main content of an image well.


\paragraph{Discussion}
Alternative supervision approaches can be conceived. 
As a specific potential alternative, one may think of directly supervising pixel-level attention map (or shift map) to provide more precise location information.
However, it is totally anecdotal how to annotate while incorporating semantic information.
Also, if we create directly retargeted images using pixel-level annotations and use them as output, we have to make images with all aspect ratios. Also, it is not desirable to simply set the results of image retargeting as a unique result. 
In this regard, we decide to use only image-level annotations as supervision, and try to resolve artifacts such as wobbles through the 1D duplicate convolution layer and structure loss.

When retargeting is performed, regions with a high value in the attention map are reduced more than areas with a low value. As in~\figref {fig:distort}-(c,d), when we look at the attention map, high values are formed in the shape of curved lines along the background, which is similar to seams to be removed in seam carving. That is, our proposed method shrinks less important parts, similar to seam carving, but uses high-level information induced by semantic understanding.

An image without dominant objects (e.g. landscape) can be regarded as a limitation because the prominent object in an image is either too small or does not appear. However, the network can still perform reasonably well because it has higher responses in textural areas in comparisons with textureless areas.



\section{Conclusion}
In this paper, we have proposed a weakly- and self-supervised deep convolutional neural network (WSSDCNN) for content-aware image retargeting. Our network produces a retargeted image directly, given an input image and a target aspect ratio. The network architecture consists of an encoder-decoder structure for the attention map generation, and the shift map is generated from 1D duplicate convolution and cumulative normalization layers. Because our network is trained in an end-to-end manner using only image-level annotations, we do not have to collect a large amount of finely annotated image data. The proposed content and structure losses ensure a retargeted image preserves important objects, and also reduce visual artifacts such as structure distortion. Experimental results demonstrate that our algorithm is superior to previous works, and shows more visually pleasing qualitative results. 

\paragraph{Acknowledgements}
We thank the anonymous reviewers for their valuable comments, and all the participants in our user study.
\clearpage
\clearpage

{\small
\bibliographystyle{ieee}
\bibliography{egbib}
}

\end{document}